\documentclass[10pt,twocolumn,letterpaper]{article}

\usepackage[pagenumbers]{cvpr} 

\usepackage{graphicx}
\usepackage{amsmath}
\usepackage{amssymb}
\usepackage{booktabs}
\usepackage{mathtools}
\usepackage{booktabs}
\usepackage{xcolor}
\usepackage{multirow}
\usepackage{microtype}
\usepackage[accsupp]{axessibility}

\usepackage[pagebackref,breaklinks,colorlinks]{hyperref}

\usepackage[capitalize]{cleveref}
\crefname{section}{Sec.}{Secs.}
\Crefname{section}{Section}{Sections}
\Crefname{table}{Table}{Tables}
\crefname{table}{Tab.}{Tabs.}

\def\cvprPaperID{3} 
\def\confName{CVPR}
\def\confYear{2022}

\begin{document}

\title{How Do You Do It? Fine-Grained Action Understanding with Pseudo-Adverbs}

\author{Hazel Doughty\\
University of Amsterdam\\
{\tt\small hazel.doughty@uva.nl}
\and
Cees G.M. Snoek\\
University of Amsterdam\\
{\tt\small cgmsnoek@uva.nl}
}
\maketitle

\begin{abstract}
\vspace{-0.8em}
We aim to understand how actions are performed and identify subtle differences, such as `fold firmly' vs. `fold gently'. To this end, we propose a method which recognizes adverbs across different actions. However, such fine-grained annotations are difficult to obtain and their long-tailed nature makes it challenging to recognize adverbs in rare action-adverb compositions. Our approach therefore uses semi-supervised learning with multiple adverb pseudo-labels to leverage videos with only action labels. 
Combined with adaptive thresholding of these pseudo-adverbs we are able to make efficient use of the available data while tackling the long-tailed distribution.
Additionally, we gather adverb annotations for three existing video retrieval datasets, which allows us to introduce the new tasks of recognizing adverbs in unseen action-adverb compositions and unseen domains. Experiments demonstrate the effectiveness of our method, which outperforms prior work in recognizing adverbs and semi-supervised works adapted for adverb recognition.
We also show how adverbs can relate fine-grained actions. 
\end{abstract}

\vspace{-1.2em}
\section{Introduction}
\vspace{-0.5em}
\label{sec:intro}



This paper aims to recognize fine-grained differences between actions such as whether a person is swimming \textit{slowly} or \textit{quickly} or cutting \textit{evenly} or \textit{unevenly}. 
Understanding how actions are performed is key to understanding the actions themselves and their outcomes. Improved perception of the action manner would allow both humans and robots  
to better imitate actions, as well as 
better discrimination between fine-grained action categories, where the difference can simply be  
how much an object moves~\cite{goyal2017something}. 
Previous works can address the question of \textit{what} is happening in a video~\cite{zhu2020comprehensive}, \textit{when} an action is happening~\cite{xia2020survey}, \textit{who} is performing an action~\cite{ye2021deep} and \textit{where} it is taking place~\cite{lowry2015visual}. However, very few works have looked at \textit{how} actions happen, as we do in this paper. 

\begin{figure}
    \centering
    \includegraphics[width=\linewidth]{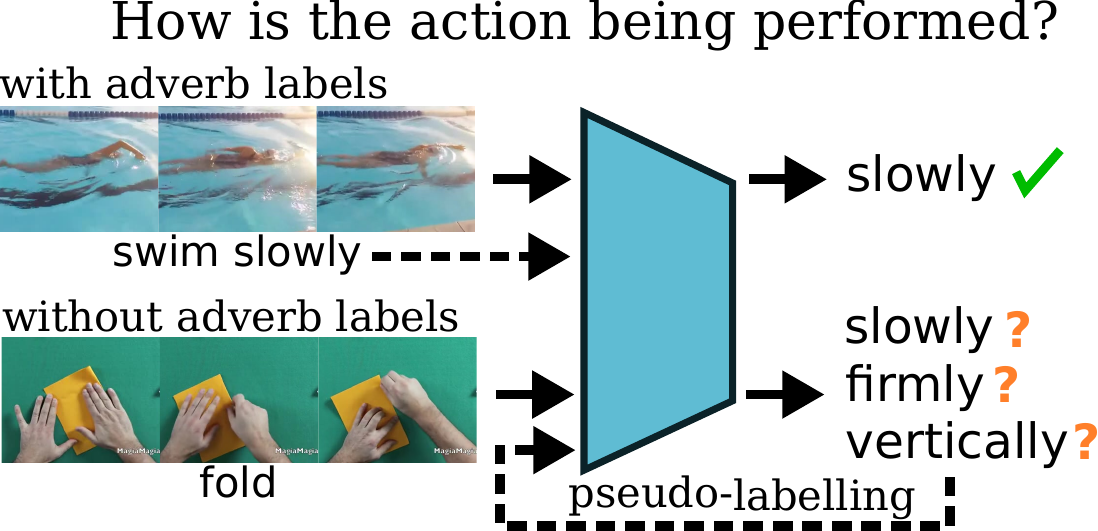}
    \vspace{-1.5em}
    \caption{We answer  
    how actions are happening by learning adverbs of 
    different actions. 
    We do this in a semi-supervised manner where we use action-only videos with multi-adverb pseudo-labeling.}
    \vspace{-1.2em}
    \label{fig:concept}
\end{figure}

In language, how an action is performed can be described with adverbs, thus we focus on recognizing such adverbs.
Two works have previously investigated adverb recognition~\cite{doughty2020action, pang2018human}. However, these works either focus on adverbs describing facial expressions and moods~\cite{pang2018human} or only studied a handful of adverbs~\cite{doughty2020action}
limiting the ways to answer ``how is the action being performed?''. This highlights a key challenge in learning adverbs and more generally fine-grained video understanding:  
the time-consuming data collection. The more subtle the differences between videos, the more difficult it is to collect a large amount of labels. 
To address these challenges and better describe how actions are performed, we scale up the number of adverbs which can be learned by utilizing videos with only action labels. 
Furthermore, multiple adverbs can co-occur and apply to the same action. We can thus better learn adverbs from videos in a semi-supervised fashion by obtaining extra adverb labels through multi-adverb pseudo-labeling 
(see Fig.~\ref{fig:concept}).

As our main contribution, we propose to reformulate the adverb recognition problem as a semi-supervised learning problem. In Sec.~\ref{sec:method}, we propose the first method for semi-supervised learning of adverbs, in which we apply multiple adverb pseudo-labels to actions and use an adaptive threshold to cope with the long-tail distribution of adverbs. In Sec.~\ref{sec:datasets},
we create several new adverb recognition benchmarks by automatically mining action-adverb pairs from the captions in existing video retrieval datasets~\cite{krishna2017dense, wang2019vatex, xu2016msr}.
Alongside this we propose two new tasks for addressing how actions happen: first recognizing adverbs in unseen compositions and second in recognizing adverbs across domains. In Sec.~\ref{sec:exp}, we demonstrate our multi-adverb pseudo-labeling approach obtains a considerable improvement over prior works in recognizing seen compositions of verbs and adverbs as well as improving generalization in these new tasks.

\section{Related Work}
\vspace{-0.5em}
We first review works focused on fine-grained understanding of actions followed by video retrieval. We then examine works which have focused specifically on adverbs. Finally, we look at semi-supervision for other vision tasks. 

\smallskip
\noindent\textbf{Fine-Grained Action Understanding. }
Recent datasets 
have focused on fine-grained 
actions~\cite{goyal2017something, li2018resound, damen2018scaling, shao2020finegym}. 
For instance, in FineGym~\cite{shao2020finegym} a model must distinguish between `salto forwards' and `salto backwards'. 
While some actions are similar, the majority of works~\cite{carreira2017quo, feichtenhofer2019slowfast, kwon2021learning, lin2019tsm, liu2021no, tran2018closer, wang2021action, zhu2018fine} model actions as distinct categories leaving the model to implicitly learn similarities.
Some works  
instead explicitly model actions as compositions of 
components, either through sub-actions~\cite{piergiovanni2018fine, piergiovanni2020differentiable} or verbs and noun combinations with~\cite{ji2020action, mettes2021object} or without~\cite{baradel2018object, damen2018scaling, jain2015objects2action, materzynska2020something, shao2020intra, zhukov2019cross} the noun's spatial location. We instead, look at fine-grained differences between actions by recognizing adverbs in combination with different verbs.

Other works recognize actions through combinations of specified attributes
~\cite{liu2011recognizing, rohrbach2016recognizing, rosenfeld2018action, zellers2017zero, zhang2021temporal}. For instance, with Temporal Query Networks, Zhang~\textit{et al.}~\cite{zhang2021temporal} propose  
to determine the correct attributes by first attending to the most relevant video parts with an attribute-focused query. The attributes studied in these works do not consider adverbs, instead they indicate the presence of an object, a person's pose or the number of repetitions of an action.

\smallskip
\noindent\textbf{Video Retrieval. }
Potentially more fine-grained than action recognition is video-text retrieval, which aims to retrieve the correct caption describing the video. 
The majority of such works create sentence-level features with 
recurrent networks~\cite{dong2019dual, hendricks2017localizing,  mithun2018learning}, learned pooling~\cite{miech2017learnable} or 
transformers~\cite{gabeur2020multi, liu2019use, yang2021taco, zhu2020actbert}. While retrieval datasets~\cite{hendricks2017localizing, krishna2017dense, lei2020tvr, oncescu2020queryd, wang2019vatex, xu2016msr} do contain adverbs, models use verbs and nouns to distinguish videos as they are more frequent~\cite{wray2019fine}. 
Rather than relying on a sentence encoding to indicate the distinctive elements of a caption, some prior works 
focus on certain parts of speech~\cite{chen2020fine, wray2019fine, xu2015jointly}. Again, the focus is on verbs and nouns, with
Wray~\textit{et al.}~\cite{wray2019fine} learning separate embeddings for each
and Chen~\textit{et al.}~\cite{chen2020fine} learning a hierarchical text encoding from verbs, nouns 
and the semantic relation between them.
We instead focus on understanding adverbs and how these apply to different verbs. We obtain new, more varied, action-adverb annotations 
from three video retrieval datasets.

\smallskip
\noindent\textbf{Adverbs. }
Some works have studied individual adverbs. For instance, Benaim~\textit{et al.}~\cite{benaim2020speednet} identify whether videos are played \textit{quickly}, Epstein~\textit{et al.}~\cite{epstein2020oops} recognize whether an event occurred \textit{accidentally} and Heidarivincheh~\textit{et al.}~\cite{heidarivincheh2018action} pinpoint when an action has finished \textit{completely}.  
There are two prior works which look at recognizing adverbs more generally. Pang~\textit{et al.}~\cite{pang2018human} propose a fully-supervised method using video, pose and expression features. 
The adverbs in 
this work focus primarily on moods and expressions such as \textit{solemnly} and \textit{excitedly}. Doughty~\textit{et al.}~\cite{doughty2020action} learn adverbs from weak supervision 
with attention locating the video segments relevant to the action. Adverbs are then learned as transformations in an embedding.
This approach is still label-hungry, requiring sufficient adverb-labeled actions for all action-adverb compositions. 
We instead utilize action-only labeled videos to recognize adverbs in both seen and unseen compositions. 
For this we introduce three new adverb datasets, significantly increasing the number of adverbs from 6 to 34 and the number of compositions from 263 to 1,550.

\begin{figure*}
    \centering
    \includegraphics[width=0.95\linewidth]{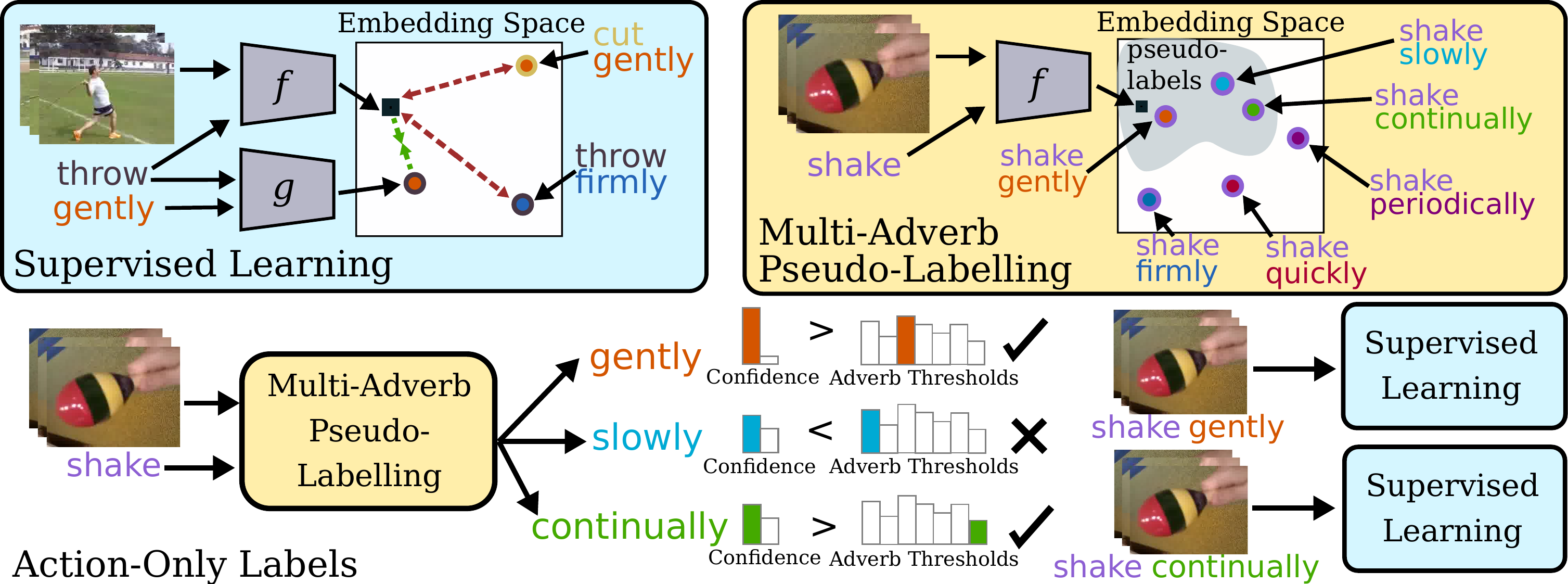}
    \vspace{-0.6em}
    \caption{Semi-supervised learning of adverbs. In the supervised case our input is the video with an action and adverb label, \textit{e.g.} throw \textit{gently}. $f$ embeds the video parts relevant to the action close to the ground-truth action-adverb text embedding from $g$.
    For videos without adverb labels we create multiple pseudo-labels by finding the most confident adverb predictions when compared to their antonym. In this example, for the action shake we obtain the pseudo-adverbs \textit{slowly}, \textit{gently} and \textit{continually}. We use per-adverb thresholds to select which of these pseudo-labels we should use. Each video is then trained with the selected pseudo-labeled adverbs as if they were in the supervised set.}
    \vspace{-0.8em}
    \label{fig:method}
\end{figure*}

\smallskip
\noindent\textbf{Semi-Supervision.}
Many strategies have been explored for semi-supervised learning such as pseudo-labeling~\cite{arazo2020pseudo, grandvalet2004semi, lee2013pseudo}, consistency regularization~\cite{bachman2014learning, berthelot2019mixmatch, singh2021semi, sohn2020fixmatch}, generative models~\cite{odena2016semi, rasmus2015semi} and fine-tuning self-supervised models~\cite{zhai2019s4l}. For instance, Lee~\cite{lee2013pseudo} proposes an efficient method for pseudo-labeling where one-hot labels are obtained for an unlabeled sample by taking the highest confidence prediction. 
Sohn~\textit{et al.} propose the consistency regularization approach FixMatch~\cite{sohn2020fixmatch},
where the loss aims to make the label predicted for two augmented versions of an image consistent.

Several works focus on semi-supervised learning for video~\cite{gavrilyuk2021motion, jing2021videossl, singh2021semi, xiong2021multiview}. TCL by Singh~\textit{et al.}~\cite{singh2021semi} maximizes the prediction similarities between different speeds of a video. 
Xiong~\textit{et al.}~\cite{xiong2021multiview} target consistency in the pseudo-labels predicted by 
RGB, optical flow and temporal gradient streams. Gavrilyuk~\textit{et al.}~\cite{gavrilyuk2021motion} also propagate pseudo-labels between modalities, but instead aim to distill motion information so downstream tasks only need the RGB modality in training.

Since these works target image, object or action recognition, they are unsuitable for adverbs. Adverbs are compositional both with actions and other adverbs and these compositions have a long-tailed distribution.  
We propose a semi-supervised approach to learn adverbs via multi-adverb pseudo-labeling and adaptive thresholding to address these challenges. We also demonstrate how our approach can improve generalization to unseen action-adverb compositions.

\section{Semi-supervised Learning of Adverbs} \label{sec:method}
\vspace{-0.5em}
Our work aims to understand how an action is being performed in a video by predicting the adverb(s) applicable to that action. An overview of our approach can be seen in Fig.~\ref{fig:method}. Labeled data can be used to learn to recognize adverbs in composition with different actions 
(Sec.~\ref{sec:supervised}). 
However, a key challenge in understanding subtle differences,
such as adverbs, is lack of labeled data. In this work we propose to better learn adverbs with semi-supervised learning by creating pseudo-adverb labels for video clips with action labels only (Sec.~\ref{sec:pseudolabel}). 
We observe that multiple adverbs can apply to the same action, therefore we propose 
to better utilize the available data 
with our multi-adverb pseudo-labeling (Sec.~\ref{sec:multilabel}). 
Another challenge is the long-tailed nature of adverbs.
We use adaptive thresholding on the adverb pseudo-labels so that our approach is effective on all adverbs, not only the most frequent (Sec.~\ref{sec:adaptive_thres}).  

\subsection{Learning Adverbs with Labeled Data}
\label{sec:supervised}
\vspace{-0.5em}
Given a video clip $x \in X$ and a label for the action of interest $a \in A$, the goal of adverb recognition is to correctly predict the adverb $\hat{m}$ which applies to action $a$. Since many adverbs are not mutually exclusive and can simultaneously apply to the same action, we particularly focus on 
distinguishing between the labeled adverb $m$ and its antonym $\mathrm{ant}(m)$. In the supervised case we learn to recognize adverbs with a labeled set of videos $S {=} \{(x, a, m)\}$.

As in prior work~\cite{doughty2020action}, we learn adverbs in a video-text embedding space as this allows actions and adverbs to be compositional. 
The goal is to embed the parts of the video relevant to the 
action close to a text representation of that action modified by the adverb. Specifically, we learn a visual embedder $f : X, A \rightarrow E$ and a textual embedder $g : A, M \rightarrow E$. We aim for $f(x,a)$ and $g(a,m)$ to be close in the embedding space $E$ and $f(x,a)$ to be far from embeddings with other actions $g(a',m)$ and with the antonym adverb $g(a,\mathrm{ant}(m))$. We do this with two triplet losses:
\begin{align}
\label{eq:lact}
    \mathcal{L}_{act}(S) = \smash[b]{\smashoperator{\sum_{(x, a, m) \in S}}} \mathrm{max}(0, &\mathrm{dist}(f(x, a), g(a, m)) - \\ \nonumber
    &\mathrm{dist}(f(x, a), g(a', m)) + \gamma_1) \\ \nonumber s.t. \ a \neq a',
\end{align}
\vspace{-1.5em}
\begin{align}
\label{eq:ladv}
    \mathcal{L}_{adv}(S) = \smash[b]{\smashoperator{\sum_{(x, a, m) \in S}}} \mathrm{max}(0, &\mathrm{dist}(f(x, a), g(a, m)) - \\ \nonumber
    &\mathrm{dist}(f(x, a), g(a, \mathrm{ant}(m))) + \gamma_2),
\end{align}
where $\mathrm{dist}$ is a distance metric and $\gamma_1$ and $\gamma_2$ are margins.

\subsection{Pseudo-Labeling Adverbs}
\label{sec:pseudolabel}
\vspace{-0.3em}
Now we consider how we can improve the learning of adverbs by utilizing video clips with only action labels. To make use of this type of data we propose to pseudo label action clips with adverbs. 
Formally, we have a video set without adverb labels $U {=} \{(x, a)\}$. For each video clip $x$ with action label $a$, we can create a single adverb pseudo-label $\widetilde{m}$ by selecting the adverb in the closest text representation as the pseudo-label. Let $d(x, a, m) {=} \mathrm{dist}(g(a,m), f(x,a))$:
\begin{equation}
\label{eq:singlelabel}
    \widetilde{m} = m_n \ \ \text{where} \ n = \mathrm{argmin}_{i} \ d(x, a, m_{i}), 
\end{equation}
where 
$m_n$ is a label indicating adverb $n$.
For the action-only videos we can then use $\widetilde{m}$ in place of $m$ in $L_{adv}$ (Equation~\ref{eq:ladv}). This gives us the overall loss:
\begin{equation}
\label{eq:overal_loss}
    \mathcal{L} = \mathcal{L}_{act}(S) + \mathcal{L}_{adv}(S) + \mathcal{L}_{act}(U) + \mathcal{L}_{adv}(U).
\end{equation}

\subsection{Multi-Adverb Pseudo-Labeling}
\label{sec:multilabel}
\vspace{-0.3em}
While actions in the supervised set $S$ are labeled with a single adverb, the majority of adverbs are not mutually exclusive, meaning multiple adverbs can apply to a single action. We thus propose multi-adverb pseudo labeling. To do this we take the top $k$ most confident adverbs and let the adverb pseudo-label $\widetilde{m}$ to be a set of pseudo-labels:
\begin{equation}
\label{eq:multilabel}
    \widetilde{m} = \{m_n\} \ \text{s.t.} \ n \in \mathrm{topk}_{i}(\mathrm{conf}(x,a,m_i)),
\end{equation}
where
\vspace{-1.2em}
\begin{equation}
\label{eq:conf}
    \mathrm{conf}(x,a,m) = \frac{e^{d(x,a,m)}}{e^{d(x, a, m)} + e^{d(x, a, \mathrm{ant}(m))}}.
\end{equation}

With this definition of $\mathrm{conf}(x, a, m)$ we take the most confident to mean the greatest relative difference between the adverb and its antonym rather than the closest adverbs.

Now we have multiple adverb pseudo labels for each of the action-only labeled videos in $U$. We optimize for each pseudo-labeled adverb, meaning the overall loss becomes:
\begin{equation}
\label{eq:overall_loss2}
    \mathcal{L} = \mathcal{L}_{act}(S) + \mathcal{L}_{adv}(S) + \sum_{\widetilde{m}}  (\mathcal{L}_{act}(U) + \mathcal{L}_{adv}(U)).
\end{equation}

\subsection{Adaptive Adverb Thresholding}
\label{sec:adaptive_thres}
\vspace{-0.3em}
The problem of recognizing adverbs is naturally long-tailed. Not only are some adverbs much more common than others, but certain compositions of actions and adverb are also more frequent. Using 
our multi-adverb pseudo-labeling we are able to make better use of the available data.
However, it has a tendency to only select the most frequent adverbs, as these are the adverbs it is most confident in predicting. 

We take inspiration from semi-supervised object detection where the long-tail is also present~\cite{li2021rethinking} and propose to use adaptive thresholding. The threshold is dynamically adjusted for each adverb $m$. Not only does this mean that the threshold is increased for the more confident adverbs so that fewer noisy pseudo-labels are used, but importantly the threshold is lowered for the adverbs with fewer confident predictions, meaning they are no longer unrepresented in the pseudo-labels. We adapt an initial threshold $\tau$ to an adverb-specific threshold $\tau_m$ as follows:
\vspace{-0.5em}
\begin{equation}
\label{eq:adaptive_threshold}
    \tau_{m} = \Bigg( \frac{\sum_{U: m \in \widetilde{m}}{\mathrm{conf}(x, a, m)}}{\frac{1}{N}\sum_U |\widetilde{m}|} \Bigg)^{\lambda} \tau,
\end{equation}
where $N$ is the number of adverbs. The sum of confidence scores for an adverb $m$, $\sum_{U: m \in \widetilde{m}}{\mathrm{conf}(x, a, m)}$, acts as an approximation of the model's overall confidence for predicting this adverb over its antonym. We then divide this by the average number of pseudo-labels per adverb. $\lambda$ is a smoothing factor which controls the amount the model focuses on underrepresented adverbs. With $\lambda{=}0$, all adverbs use the original threshold $\tau$. The adverb-specific threshold $\tau_m$ is applied to filter the available pseudo-labels, so that only the pseudo labels with $\mathrm{conf}(x, a, m) > \tau_m$ for $m \in \widetilde{m}$ are used.

\section{Adverb Datasets and Tasks} \label{sec:datasets}
\vspace{-0.3em}
We evaluate our approach on \textbf{HowTo100M Adverbs}~\cite{doughty2020action} which mined adverbs from  
83 tasks in HowTo100M~\cite{miech2019howto100m}. Since the annotations were obtained from automatically transcribed narrations of instructional videos, 
they are noisy; $\sim$44\% of the annotated action-adverb pairs are not visible in the video clip. The dataset contains 5,824 clips annotated with action-adverb pairs from 72 verbs and 6 adverbs. A clear limitation of this dataset is the small number of adverbs it contains, we thus create three new adverb datasets from existing video retrieval datasets~\cite{wang2019vatex, xu2016msr, krishna2017dense}: VATEX Adverbs, MSR-VTT Adverbs and ActivityNet Adverbs. These contain less noise and a greater variety of adverbs.

\begin{table*}[t]
    \centering
    \resizebox{0.9\linewidth}{!}{%
    \begin{tabular}{lrrrrrrccc}
    \toprule
    & \multicolumn{4}{c}{\textbf{Adverbs \& Actions}}  & \multicolumn{2}{c}{\textbf{Videos}} & \multicolumn{3}{c}{\textbf{Tasks}}\\
    \cmidrule(lr){2-5} \cmidrule(lr){6-7} \cmidrule(lr){8-10}
    \textbf{Dataset} & Adverbs & Actions & Pairs & Accuracy & Clips & Length (s) & Seen & Unseen & Domain\\
         \midrule
         HowTo100M Adverbs~\cite{doughty2020action} & 6 & 72 & 263 & 44.0\% & 5,824 & 20.0 & $\checkmark$ & - & - \\
         \midrule
         VATEX Adverbs & 34 & 135 & 1,550 & 93.5\% & 14,617 & 10.0 & $\checkmark$ & $\checkmark$  & Source \\
         MSR-VTT Adverbs & 18 & 106 & 464 & 91.0\% & 1,824 & 15.7 & $\checkmark$ & - & 
         Target \\
         ActivityNet Adverbs & 20 & 114 & 643 & 89.0\% & 3,099 & 37.3 & $\checkmark$ & - &
         Target \\ 
         \bottomrule
    \end{tabular}
    }
    \vspace{-0.8em}
    \caption{Our three newly proposed 
    adverb datasets have more adverbs, actions, unique pairs and higher annotation accuracy than HowTo100M Adverbs~\cite{doughty2020action} and also allow us to study recognition of adverbs in unseen action-adverb compositions and unseen domains.}
    \vspace{-0.8em}
    \label{tab:dataset_comp}
\end{table*}

\subsection{Adverb Annotations from Video Captions}
\vspace{-0.3em}
We extract verb-adverb annotations for videos in existing video-text datasets to obtain three new adverb datasets. From available 
datasets~\cite{gao2017tall, hendricks2017localizing, huang2020movienet, jang2017tgif, krishna2017dense, lei2019tvqa+, lei2020tvr, lei2020vlep, liu2020violin, oncescu2020queryd, rohrbach2014coherent, sadhu2021visual,  wang2019vatex, xu2016msr, zhou2018towards} we find VATEX~\cite{wang2019vatex}, ActivityNet Captions~\cite{krishna2017dense} and MSR-VTT~\cite{xu2016msr} contain the best variety of adverbs with sufficient instances. 
VATEX consists of 35k 10 second video clips, each with 10 English captions, resulting in a total of 260k captions. In MSR-VTT each clip is 10-30 seconds and has 20 captions giving a total of 10k clips and 200k captions. ActivityNet Captions contains 20k videos with an average of 3.65 temporally localized sentences per video, resulting in a total of 100k clips and matching captions. 
Each dataset uses YouTube videos, thus some videos are no longer available. 
At the time of collection we obtained: 32,161 video clips for VATEX, 7,511 for ActivityNet and 5,197 for MSR-VTT. 
\begin{figure}
    \centering
    \includegraphics[width=\linewidth]{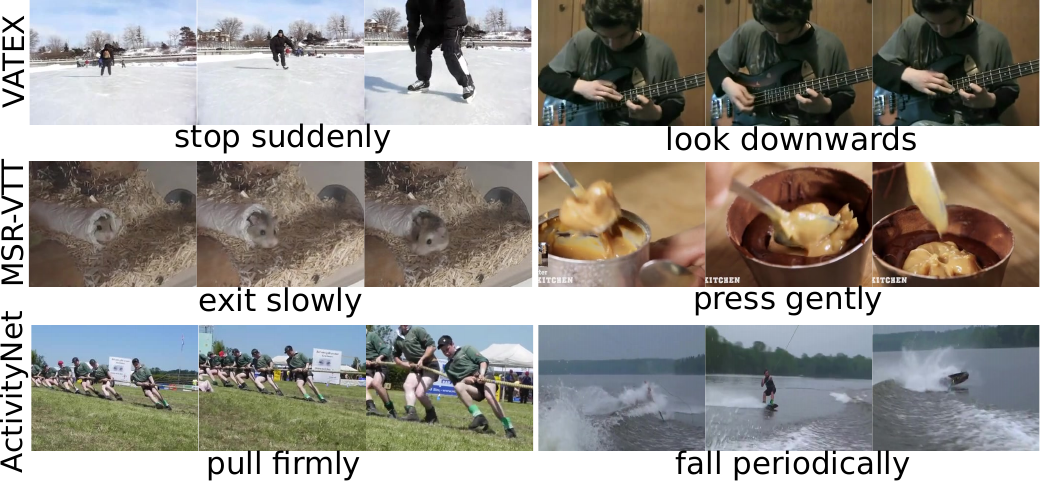}
    \vspace{-1.8em}
    \caption{Example video clips and  action-adverb annotations.}
    \vspace{-0.8em}
    \label{fig:dataset_eg}
\end{figure}

\smallskip
\noindent\textbf{Extracting Adverb Annotations. }
To extract adverb annotations from the captions in these datasets we search for adverbs and their corresponding verbs. We use SpaCy's English core web model where RoBERTa~\cite{liu2019roberta} performs Part-of-Speech tagging and dependency parsing on each caption. We search for verbs which have adverbs as children, excluding any verbs with a negative dependency to another word. We filter out non-visual adverbs,  adverbs whose antonym doesn't appear and adverbs which appear less than 10 times or only appear in combination with a single action. The resulting verbs and adverbs from the three datasets are manually clustered, starting with the clusters from~\cite{doughty2020action}. This process forms 137 verb clusters and 34 adverb clusters in 17 adverb-antonym pairs.  Fig.~\ref{fig:dataset_eg} shows examples of the video clips alongside the discovered action-adverb pairs.

\begin{figure}
    \centering
    \includegraphics[width=\linewidth]{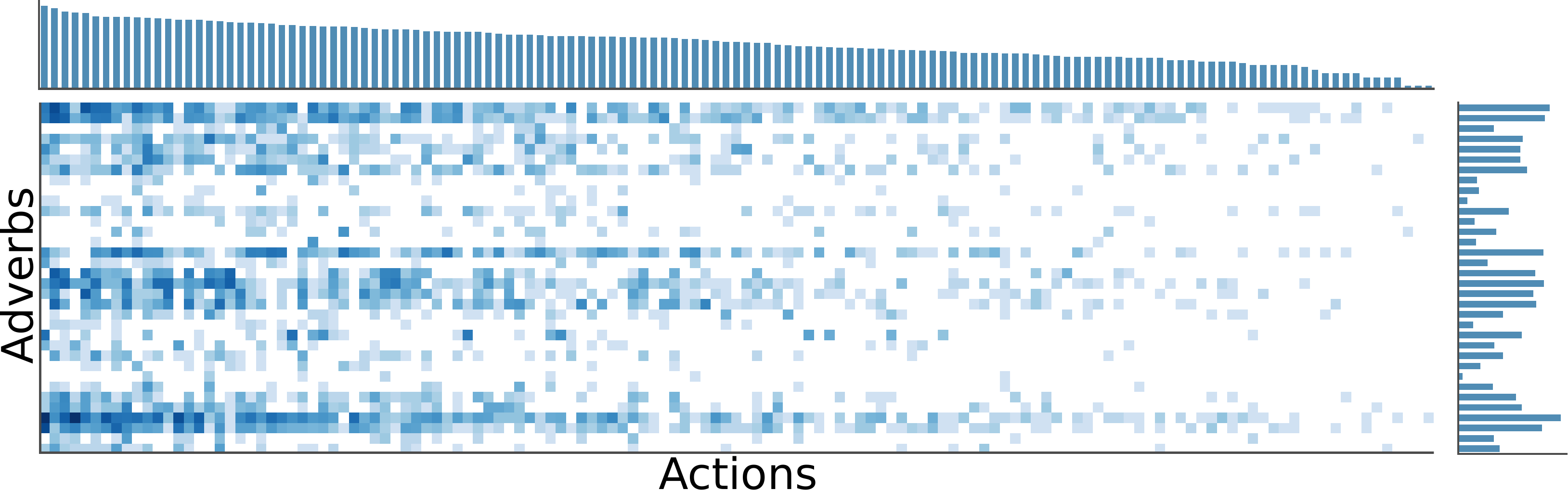}
    \vspace{-1.7em}
    \caption{Distribution of action-adverb pairs in VATEX Adverbs shown on a log-scale. The adverbs and actions labels are long tailed as are their compositions. Labeled version in supplementary.}
    \vspace{-0.8em}
    \label{fig:vatex_dist}
\end{figure}

\smallskip
\noindent\textbf{Adverb Datasets. }
This results in three adverb datasets: \textbf{VATEX Adverbs}, \textbf{ActivityNet Adverbs} and \textbf{MSR-VTT Adverbs} which we make publicly available alongside the code\footnote{\url{https://github.com/hazeld/PseudoAdverbs}}. Table~\ref{tab:dataset_comp} shows statistics of each. VATEX Adverbs is the largest with 34 adverbs appearing across 135 actions to form 1,550 unique action-adverb pairs. The distribution of actions, adverbs and their compositions are heavily long-tailed (see Fig.~\ref{fig:vatex_dist}).  
Each dataset considers many more adverbs than the existing HowTo100M Adverbs 
which contains only 6. 
We measure the quality of each dataset's annotations with a 200 video sample. Since the new datasets come from human written captions, where a person has explicitly chosen the adverb to describe the action, the annotations are much less noisy than HowTo100M Adverbs.

\subsection{Adverb Recognition Tasks}
\vspace{-0.3em}
With these datasets, 
as in prior work~\cite{doughty2020action}, we want to learn to recognize adverbs in previously seen action-adverb compositions. We additionally propose two new adverb recognition tasks: first in
unseen compositions and second in unseen domains. 
We explain each 
below.

\smallskip
\noindent\textbf{Task I: Seen Compositions. }
Adverbs and actions are compositional, an adverb $m \in M$ can apply to many different actions $a\in A$.
Assume we have a set of action-adverb compositions $(a, m) \in C$. When recognizing adverbs in seen compositions, all compositions in the test set are present in the labeled training set, \ie $C_{test} \subseteq C_{labeled}$. This tests whether the model can successfully compose and recognize adverbs across various actions.
For this evaluation we use our newly proposed VATEX Adverbs as well as HowTo100M Adverbs~\cite{doughty2020action}. We partition VATEX Adverbs into train and test following the original train and test split. This gives 11,782 video clips in training and 2,835 in testing over the 34 adverbs. 
HowTo100M-Adverbs contains 6 adverbs and consists of 5,475 video clips in training and 349 in testing.

\smallskip
\noindent\textbf{Task II: Unseen Compositions.}
To fully capture the compositional nature of actions and adverbs it is necessary for a model to generalize beyond seen compositions. 
We thus propose to recognize adverbs in
unseen compositions, \ie $C_{test} \cap C_{labeled} {=} \emptyset$.
We focus on VATEX Adverbs for this since it has the most action-adverb pairs. We partition the pairs into two disjoint sets.
 For each action, both the pair with an adverb and its antonym are in the same set. Each set contains 50\% of the pairs and every action and adverb is present in both sets. We take one split for training and further partition the second split, using 
 half the instances of each pair as the test set and half as the action-only set.

\smallskip
\noindent\textbf{Task III: Unseen Domains. }
Since a key challenge of fine-grained video understanding is the collection of labeled data, we cannot assume to have labels in every domain where we wish to recognize adverbs. We thus propose to test the transferability of learned adverbs to new domains. Here our labeled data $S$ comes from a domain $D_{S}$ while our test set and action-only labeled data $U$ come from a distinct domain $D_{U} {\neq} D_{S}$.
We use VATEX Adverbs as the source and MSR-VTT Adverbs and ActivityNet Adverbs as targets. We partition both targets in two 50\% splits, one for testing and the other as action-only labeled training data. 

\section{Experiments}\label{sec:exp}
\vspace{-0.3em}
We first describe the implementation details of our method and the evaluation metric used. We then analyze the contribution of our model's components and compare to semi-supervised baselines for recognizing adverbs in seen compositions. Finally, we evaluate our approach for recognizing adverbs in unseen compositions and unseen domains.

\begin{figure*}
    \centering
\includegraphics[width=\linewidth]{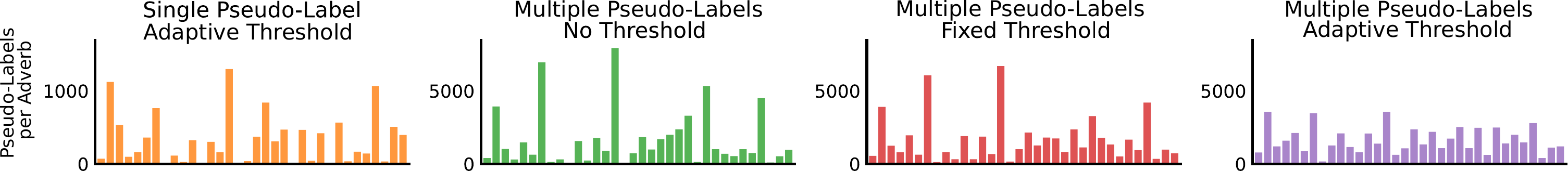}
   \vspace{-2em}
    \caption{Distribution of adverb pseudo labels over all videos. Each bar indicates the number of videos pseudo-labeled with a particular adverb. With multi-adverb pseudo-labeling and adaptive thresholding in our model (purple), pseudo-labels are better distributed among the possible adverbs. With either single-adverb pseudo-labels (yellow) or other types of thresholding (green and red) the pseudo-labels reflect the long-tail distribution of ground-truth labels.}
       \vspace{-1.5em}
    \label{fig:pseudo_label_dist}
\end{figure*}

\smallskip
\noindent\textbf{Implementation Details. }
All videos are sampled at 25fps and scaled to 256px in height. Each video is divided into 1-second segments with one 16-frame snippet extracted per segment. We use a frozen I3D network as the backbone, one for RGB and one for optical flow. The output of the global pooling layer for each modality  
is concatenated to create a $T {\times} 2048D$ feature, where $T$ is the length of the video clip in seconds. The video embedder $f$ uses transformer-style attention to locate the relevant video parts with the $T$ video features as the keys and the action as the query. The text embedder $g$ uses GloVe embeddings to represent the actions and learns adverbs as linear transformations on action embeddings. See~\cite{doughty2020action} for more details. Optimization is done with Adam~\cite{kingma2014adam}. Models are trained with a supervised batch size of 128 and learning rate of $10^{-4}$ for 1000 epochs. As in~\cite{doughty2020action} we introduce the adverbs after the 200th epoch, until that moment we train $g$ as an action embedder.   
In experiments without thresholding, we reduce noise by letting the adverb representations train for 100 epochs before introducing pseudo-labels.
The ratio of adverb-labeled to action-only labeled samples in a batch is the same as the 
total ratio.
Unless otherwise specified, we set the triplet loss margins to $\gamma_1{=}\gamma_2{=}1$ (Eq.~\ref{eq:lact}, \ref{eq:ladv}), the maximum pseudo-labels per video to $k{=}5$ (Eq.~\ref{eq:multilabel}), the base threshold to $\tau {=} 0.6$ (Eq.~\ref{eq:adaptive_threshold}) and the smoothing factor to $\lambda{=}0.1$ (Eq.~\ref{eq:adaptive_threshold}).
 
 \smallskip
\noindent\textbf{Evaluation Metric.}
We use adverb-antonym binary classification accuracy from~\cite{doughty2020action}. 
That is the accuracy when considering the ground-truth adverb vs. its antonym. This suits the available ground-truth labels since they indicate a single adverb, while multiple adverbs may apply to an action.
As the distributions of adverbs in our new datasets 
are long-tailed, we report the average accuracy over adverbs for these, rather than average over videos.

\begin{figure}
    \centering
    \includegraphics[width=0.95\linewidth]{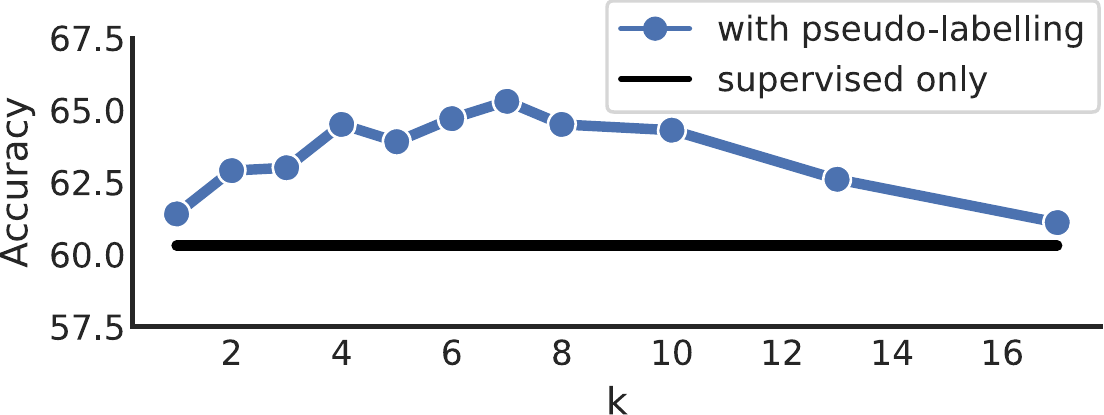}
    \vspace{-1em}
    \caption{Performance when changing $k$, the max pseudo-labels per video. Using multi-adverb pseudo-labeling improves performance.} 
    \vspace{-0.8em}
    \label{fig:k}
\end{figure}

\subsection{Ablation Study}
    \vspace{-0.3em}
We first perform several ablation studies evaluating the effect of each of the proposed model components. For these we recognize adverbs in seen compositions with  
VATEX Adverbs since this has the greatest variety of  
adverbs. Experiments are performed with 5\% of the training set as the labeled set and the remainder as the action-only labeled set. 

\smallskip
\noindent\textbf{Multi-Adverb Pseudo-Labeling.}
Fig.~\ref{fig:k} shows the effect of $k$, the maximum pseudo-labeled adverbs per video. Using multiple pseudo-labels ($k{>}1$) offers a large advantage over supervised-only learning and semi-supervised learning with a single pseudo-label ($k{=}1$). 
The best performance is with $k{=}7$, although all values in $4 {\geq} k {\geq} 10$ are good. With many different adverbs applying to an action ($k{\geq}13$) the performance drops, since this many adverbs rarely co-occur, although this is still better than supervised only learning.

Using multi-adverb pseudo-labeling lets us to make more efficient use of the data at our disposal as each video clip is used to learn multiple adverbs. It also encourages exploration of less frequently labeled adverbs, which we show in Fig.~\ref{fig:pseudo_label_dist}.
 With a single adverb pseudo-label (yellow), the overall distribution of pseudo-labels is highly imbalanced and mimics the long-tail distribution of the ground-truth. There are even 5 adverbs which do not have any pseudo-labels.
Our multi-adverb pseudo-labeling (purple) reduces this bias, 
with pseudo-labels better distributed across possible adverbs.

\begin{table}[]
\vspace{-0.7em}
\begin{minipage}[b]{0.48\linewidth}\centering

    \resizebox{0.85\linewidth}{!}{\begin{tabular}{lc}
    \toprule
         Method &  Acc. \\
         \midrule
         Closest & 61.7\\
         Max Difference & 63.9\\
         \bottomrule
    \end{tabular}}
    \vspace{-0.8em}
    \caption*{(a)}
    \end{minipage}
\begin{minipage}[b]{0.48\linewidth}\centering
    \resizebox{0.8\linewidth}{!}{%
    \begin{tabular}{lc}
    \toprule
         Thresholding &  Acc. \\
         \midrule
         None & 61.1\\
         Fixed & 61.4\\
         Adaptive & 63.9\\
         \bottomrule
    \end{tabular}
    }
    \vspace{-0.8em}
    \caption*{(b)}

\end{minipage}\hfill
\vspace{-1.2em}
\caption{(a) Pseudo-label selection. Considering antonyms with max difference is better than using the closest adverbs. (b) Type of thresholding. Adaptive thresholding gives better pseudo-labels.}
\vspace{-0.6em}
\label{tab:label_and_thres}
\end{table}

\begin{figure}
    \centering
    \includegraphics[width=0.95\linewidth]{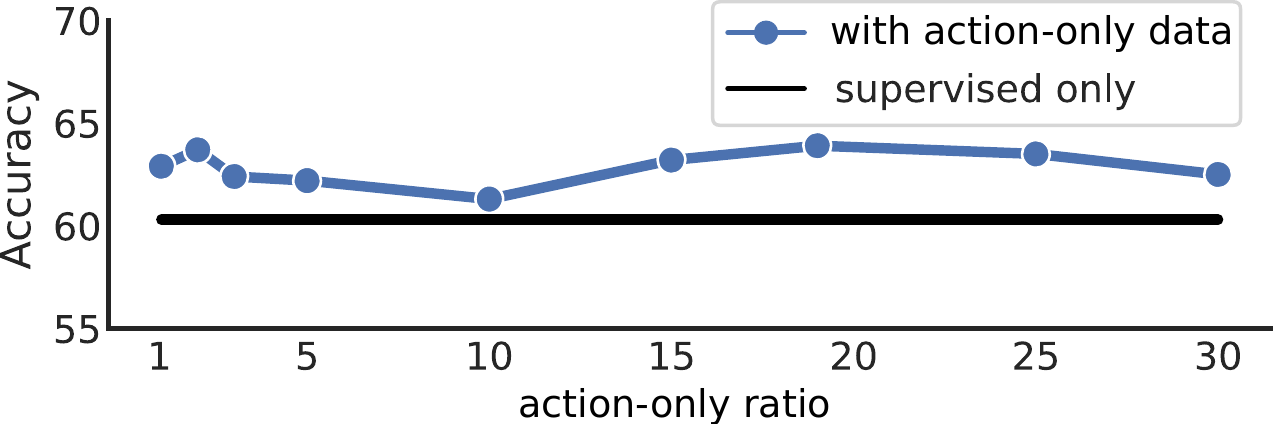}
    \vspace{-0.8em}
    \caption{Effect of the ratio of adverb-labeled to action-only data. Any ratio of action-only data improves over using just labeled data.} 
    \vspace{-0.8em}
    \label{fig:unlabelled_ratio}
\end{figure}

\smallskip
\noindent\textbf{Pseudo-Label Selection.}
A standard approach to pseudo-labeling in an 
 embedding space would be to take the closest embeddings as the pseudo-label(s).
We instead take the adverbs with the greatest difference between the embedded video's proximity to the adverb modified action and the antonym modified action. We compare these approaches in Table~\ref{tab:label_and_thres}a, which shows our approach improves the result.

\begin{table*}[]
    \centering
    \resizebox{0.8\textwidth}{!}{\begin{tabular}{lccccccccccccc}
    \toprule
        & \multicolumn{6}{c}{VATEX Adverbs} & &  \multicolumn{6}{c}{HowTo100M Adverbs}\\
        \cmidrule{2-7} \cmidrule{9-14}
         Method &  1\% & 2\% & 5\% & 10\% & 20\% & Av. & & 1\% & 2\% & 5\% & 10\% & 20\%& Av.\\
         \midrule
         Supervised only
         & 54.0 & 54.5 & 60.3 & 64.7 & 64.2 & 59.5 & & 67.3 & \textbf{68.5} & 67.9 & 73.4 & 74.8 & 70.4\\
         Pseudo-Label
         & 55.1 & 54.4 & 60.4 & 63.5 & 64.1 & 59.5 & & \textbf{69.3} & 66.5 & 67.3 & 74.5 & 70.5 & 69.6\\
         FixMatch
         & \textbf{55.4} & 52.3 & 61.2 & 62.8 & 64.8 & 59.3 && 68.2 & 67.9 & 67.3 & 74.5 & 75.9& 70.7\\
         TCL
         & 51.6 & \textbf{56.6} & 58.3 & 58.0 & 64.8 & 57.9 & & 67.6 &  65.9 & 68.2 & 74.3 & 76.2 & 70.4\\
         \midrule
         Ours & 55.0 & \textbf{56.6} & \textbf{63.9} & \textbf{65.3} & \textbf{67.5} & \textbf{61.7} & & 67.0 & 66.8 & \textbf{69.9} & \textbf{77.1} & \textbf{79.1} & \textbf{72.0}\\
         \bottomrule
    \end{tabular}
    }
        \vspace{-0.7em}
    \caption{\textbf{Seen Compositions}. When using $\geq$5\% of the labeled training data our method outperforms outperforms semi-supervised baselines for recognition of adverbs in previously seen action-adverb compositions.}
        \vspace{-0.5em}
    \label{tab:vatex_seen}
\end{table*}

  \begin{figure*}
     \centering
     \includegraphics[width=\linewidth]{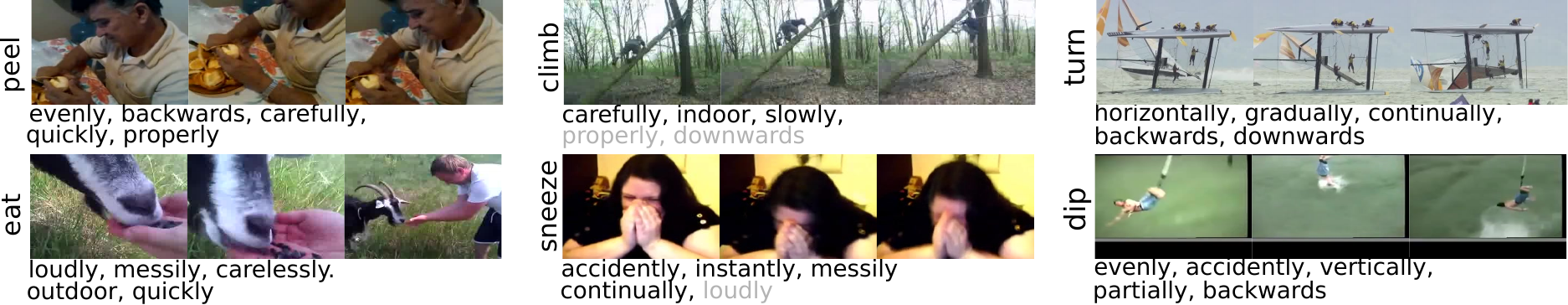}
        \vspace{-1.8em}
     \caption{Example pseudo-labels from our proposed multi-adverb pseudo-labeling for the indicated action. Pseudo-labels below their adverb threshold are shown in grey. Our method can successfully identify multiple relevant adverbs for each video (left column), can use the adaptive thresholding to ignore incorrect or unnecessary pseudo-labels (middle), but can struggle to be accurate when actions co-occur (\textit{downwards}, top right) and has no notion of situations where an adverb is infeasible (\textit{backwards}, bottom right)}
        \vspace{-0.8em}
     \label{fig:qual_res}
 \end{figure*}

\smallskip
\noindent\textbf{Adaptive Thresholding.}
Table~\ref{tab:label_and_thres}b compares the adaptive thresholding we use to no thresholding and a fixed threshold for all adverbs. The adaptive thresholding improves the result by 2.5\% over 
fixed thresholding, which 
has little impact itself. 
With fixed thresholding, once the most common adverbs have exceeded this threshold the model will pseudo-label all actions with these adverbs, ignoring rarer adverbs. 
The adaptive thresholding allows the pseudo-label selection to be more balanced (as shown in 
Fig.~\ref{fig:pseudo_label_dist}).  

\smallskip
\noindent\textbf{Ratio of Action-Only Data.}
We test the effect of the ratio of adverb-labeled to action-only labeled videos 
in Fig.~\ref{fig:unlabelled_ratio}. This shows training with any amount of action-only data gives better performance. We observe two peaks in Fig.~\ref{fig:unlabelled_ratio}. 
With an action-only ratio ${\geq}15$ 
the model is able to see all available action-only data in each epoch of the labeled data. This allows better learning of the rarer adverb-action compositions. With 2 times
the amount of action-only data the model is less likely to overfit to any noisy pseudo-labels early in training.

\subsection{Task I: Seen Compositions}
\vspace{-0.3em}
\label{sec:seen_comp}
We test our method's suitability for recognizing adverbs in previously seen action-adverb compositions as in prior work~\cite{doughty2020action}.
Thus we use HowTo100M Adverbs~\cite{doughty2020action} as well as VATEX Adverbs. 
For both datasets we test our approach with different amounts of labeled data used in training: 1\%, 2\%, 5\%, 10\% and 20\%. 
The remaining training data is used in the action-only labeled set.

We compare to the \textbf{supervised-only} adverb recognition approach Action Modifiers~\cite{doughty2020action} on which our approach is based. This learns from only the adverb-labeled data, not the action-only labeled set. We also compare to several semi-supervised approaches: Pseudo-Label~\cite{lee2013pseudo}, FixMatch~\cite{sohn2020fixmatch} and TCL~\cite{singh2021semi}, which we adapt for adverb recognition by combining them with Action Modifiers. This allows a fair comparison as the backbone and adverb representations are the same in all methods. \textbf{Pseudo-Label} 
simply takes the most confident prediction of a data sample to be the pseudo label. \textbf{FixMatch} 
obtains pseudo-labels from weak augmentations of the input data. 
Strongly augmented versions are then trained to predict the same pseudo-label. This also uses fixed thresholding. 
Instead of the image augmentations used in FixMatch, \textbf{TCL} uses the video speed.
It also optimizes agreement between the predictions for all classes rather than a single pseudo-label.
Full implementation details  
can be found in supplementary.  
 
Results are presented in Table~\ref{tab:vatex_seen}. For VATEX Adverbs our approach outperforms or obtains competitive results over all baselines for all percentages of labeled data used. On HowTo100M Adverbs our approach outperforms baselines  
for the 5\%, 10\% and 20\% labeled data settings. Our multi-adverb pseudo-labeling has more impact on VATEX Adverbs since it contains more adverbs.
The improvement is also greater when using $\geq 5$\% labeled data. With fewer labels
each adverb is seen in fewer 
situations, meaning the pseudo-labels become more noisy.
However, there is still room for further improvement over our method as using 100\% adverb-labeled data obtains 73.9\% on VATEX Adverbs and 80.8\% on HowTo100M Adverbs.
We observe that TCL often performs worse than other approaches, despite being designed for 
video. 
 This is because TCL encourages invariance to speed which affects adverbs such as 
 \textit{quickly} and \textit{slowly}.
Each of the semi-supervised baselines are comparable overall to the supervised-only method, this highlights the importance of our proposed multi-label pseudo-labelling and adaptive thresholding. Without these elements 
models are more biased to the particular action-adverb compositions.  
 
 We show examples of our multi-adverb pseudo-labeling in Fig.~\ref{fig:qual_res}. Our method provides multiple relevant adverb pseudo-labels for each video. 
 The model is able to use the adaptive thresholding to exclude incorrect predictions (climb \textit{downwards}) or 
 frequent compositions (climb \textit{properly} and sneeze \textit{loudly}). 
 There are still noisy pseudo-labels such as climb \textit{indoor} and dip \textit{evenly}. There are also cases where the adverbs make no sense 
 in the context of the action, \textit{e.g.} dip \textit{backwards}. 
 The incorrect prediction turn \textit{downwards} highlights a challenge of 
 adverb datasets, where there 
 can be multiple actions occurring at the same time. Here \textit{downwards} refers to the people falling, rather than boat turning. 

\begin{table}[]
    \centering
    \resizebox{0.6\linewidth}{!}{\begin{tabular}{lc}
    \toprule
         Method &  Accuracy\\
         \midrule
         Supervised only
         & 52.2\\
         Ours & 56.1 \\
         \midrule 
         \color{gray}{Training with full labels} & \color{gray}{65.1} \\
         \bottomrule
    \end{tabular}}
        \vspace{-0.8em}
    \caption{\textbf{Unseen compositions} in VATEX Adverbs. Our method improves generalization to unseen action-adverb compositions.} 
        \vspace{-0.5em}
    \label{tab:unseen}
\end{table}

\subsection{Task II: Unseen Compositions}
\label{sec:unseen_comp}
\vspace{-0.3em}
We investigate whether our 
method can improve recognition of adverbs in previously unseen action-adverb pairs.
We compare to supervised only 
Action Modifiers~\cite{doughty2020action}.
Table~\ref{tab:unseen} shows that our method improves performance by $\sim$4\%. 
The adaptive thresholding is key. Without it the pseudo-labels primarily consist of previously seen adverbs compositions. 
However, there is much potential for future work as highlighted by the gap between our method and training with all compositions seen.
Generalizing to unseen action-adverb composition is necessary since it is infeasible to acquire sufficient labeled data for every possible composition.

\subsection{Task III: Unseen Domains}
 \vspace{-0.3em}
In 
Table~\ref{tab:domain} we test 
whether our pseudo-labeling approach can improve transfer to new domains. 
We compare our approach to training with only the source data, \textit{i.e.}  VATEX Adverbs, as well as the Pseudo-Label~\cite{lee2013pseudo} baseline. 
Our method outperforms the Pseudo-Label approach for MSR-VTT Adverbs and gives a $\sim$2\% gain over using only source domain videos. On ActivityNet Adverbs all three approaches are comparable, 
as the gap to this dataset is larger both in terms of action and adverb appearance and action length.
Table~\ref{tab:domain} also shows the upper bounds when target data is used in training. 
The gap between our model's performance and source+target is relatively small, meaning adverb representations do transfer well between actions in different domains, however there is still  much potential for improvement in the adverb representation itself. 
This is a more realistic setting to evaluate adverb representations, since labeled data is scarce. Transferring adverb representation to new domains is key to applications such as recognizing anomalous occurrences of an action or whether someone is following a recipe well.

\subsection{Describing Relationships Between Actions} 
\begin{figure}
    \centering
    \includegraphics[width=\linewidth]{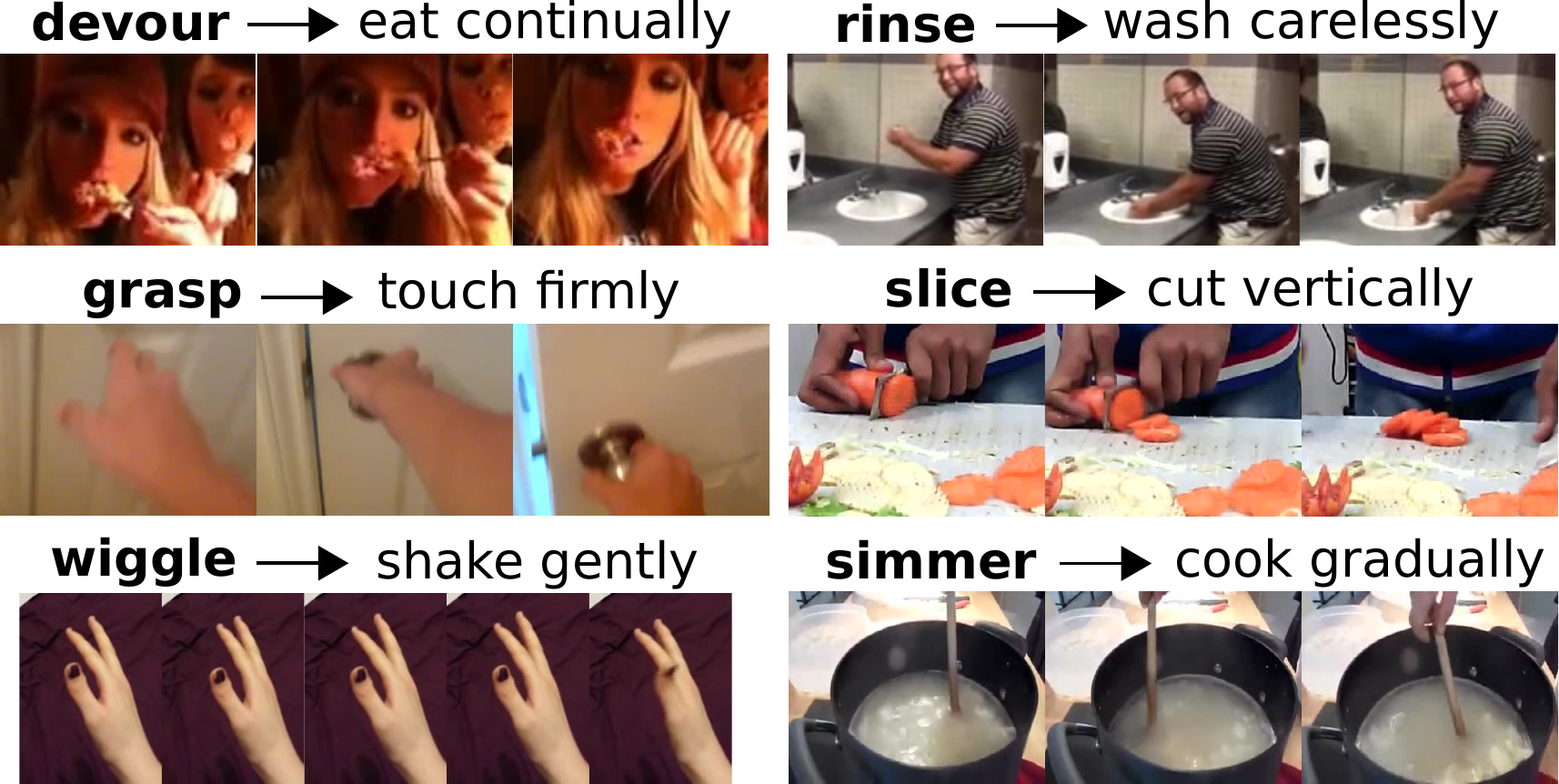}
       \vspace{-1.8em}
    \caption{We use the learned video-text embedding to identify zero-shot actions by compositions of adverbs and seen actions. We show each  zero-shot action in bold alongside the closest action-adverb pair in the embedding space and one of the closest videos.}
       \vspace{-0.8em}
    \label{fig:bonus_task}
\end{figure}

   \vspace{-0.3em}
We foresee many applications of adverbs in video understanding, such as in dense video captioning~\cite{krishna2017dense}, describing and detecting anomalies~\cite{hinami2017joint}, determining a person's skill~\cite{doughty2018s} and procedure planning~\cite{chang2020procedure}. Here we demonstrate qualitatively how adverbs can be used to identify fine-grained zero-shot actions by describing the relationship between these unseen actions and those previously seen. Fig.~\ref{fig:bonus_task} shows examples of such actions. In each case the zero-shot action can be described by applying an adverb to a known action.

\begin{table}[]
    \centering
        \resizebox{\linewidth}{!}{
        \begin{tabular}{lcc}
    \toprule
         Method &  MSR-VTT Adverbs& ActivityNet Adverbs\\
         \midrule
         Source only & 62.9 & 67.2 \\
         Pseudo-Label 
         & 63.9 & 66.4 \\
         Ours & 65.0 & 66.6\\ 
         \midrule
         \color{gray}{Source + Target} & \color{gray}{67.5} & \color{gray}{71.6}\\
         \color{gray}{Target only} & \color{gray}{70.5} & \color{gray}{71.8}\\
         \bottomrule
    \end{tabular}}
       \vspace{-0.8em}
    \caption{Transfer to \textbf{unseen domains} from VATEX-Adverbs.  
    Our method aids generalization to similar domains (MSR-VTT Adverbs), but struggles with larger shifts (ActivityNet Adverbs).}
       \vspace{-0.5em}
    \label{tab:domain}
\end{table}

\section{Discussion}
   \vspace{-0.3em}
\noindent\textbf{Limitations. }
Our method has several limitations. Firstly, our model has no concept of infeasible combinations of action and adverbs and can be confounded by co-occuring actions where different adverbs apply. It also struggles when an adverb is labeled in few contexts. While our method can aid generalization to unseen action-adverb compositions and unseen domains, there is still far to go in these areas. 

\noindent\textbf{Potential Negative Impact. } All datasets in this paper are sourced from YouTube and therefore the subjects and activities contained within are not representative of the diversity in global society. Thus our trained models will contain biases. 

\noindent\textbf{Conclusions. }This paper presents a semi-supervised method to recognize adverbs of actions. This allows us to understand how an action is being performed and understand fine-grained differences between actions. We propose multi-adverb pseudo-labeling to make use of videos with action-only labels. To cope with the long-tail distribution of adverbs and their action compositions our method also makes use of adaptive thresholding. We propose three new adverb datasets which allow us to evaluate how well our method recognizes adverbs in seen action-adverb compositions as well as unseen compositions and unseen domains. Results demonstrate our method improves performance in all three tasks. 

\noindent\textbf{Acknowledgements.} Work is part of the Real-Time Video Surveillance Search project (18038), which is partly financed by the Dutch Research Council (NWO) domain Applied and Engineering/Sciences (TTW).

{\small
\bibliographystyle{ieee_fullname}
\bibliography{egbib}
}

\section*{Supplementary Material}

\appendix
\section{Per Adverb Results}
\begin{figure}[h]
    \centering
    \includegraphics[width=\linewidth]{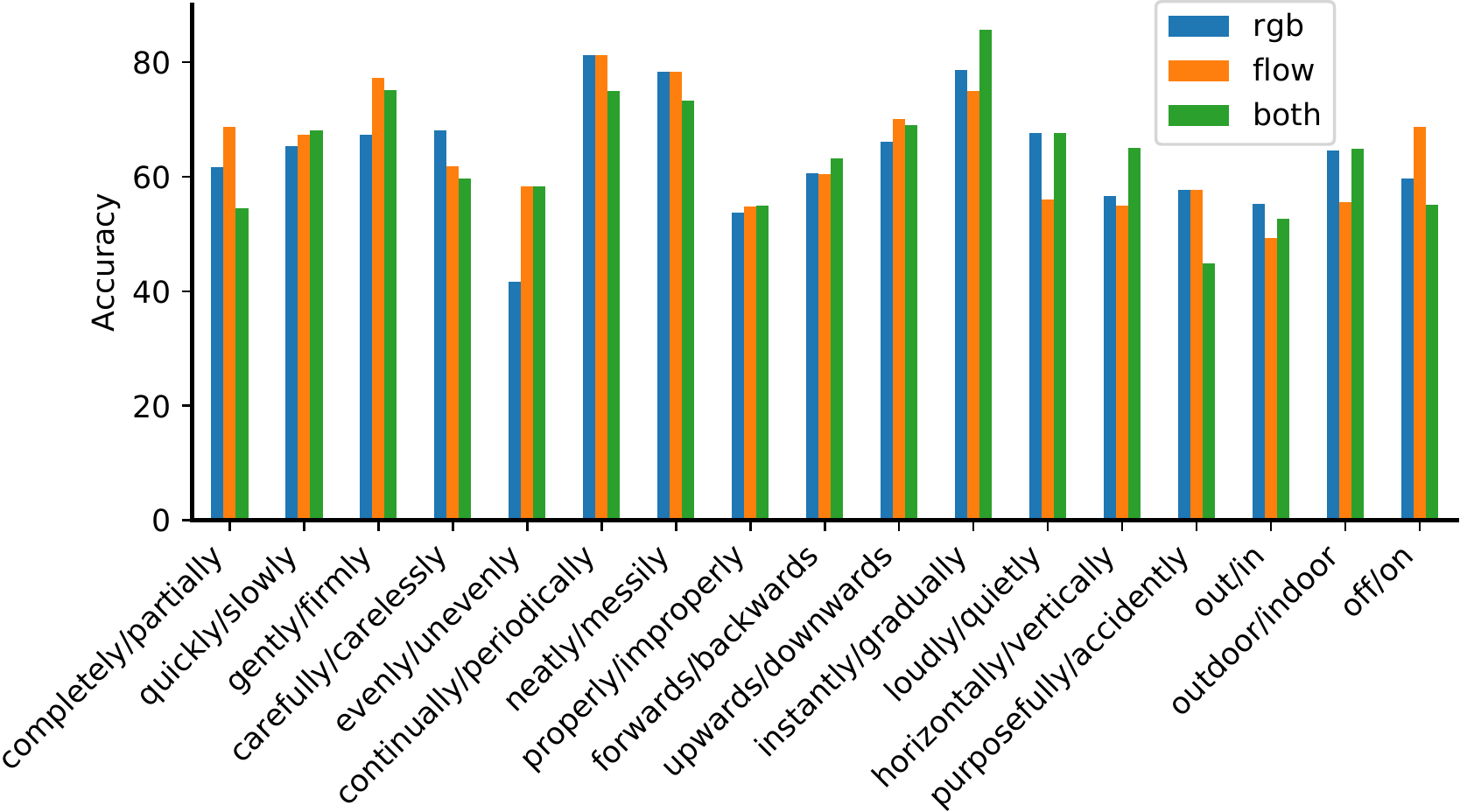}
    \vspace{-1.3em}
    \caption{Results of each modality for per adverb-antonym pair.}
       \vspace{-0.8em}
    \label{fig:modalities}
\end{figure}

\vspace{-0.3em}
We show results for each adverb-antonym pair in Fig.~\ref{fig:modalities}. We first note that our model is capable of learning each of these adverbs since all pairs perform above random performance which is 50\%. Our model performs best on `instantly/gradually', `continually/periodically' and `neatly/messily', despite the high imbalance between the number of instances in this adverb-antonym pairs. The most challenging adverb-antonym pairs are `evenly/unevenly', `properly/improperly' and `purposefully/accidentally', since the latter adverb in each pair has very few labelled samples to learn from. While our multi-adverb pseudo-labelling learns well from imbalanced data, generalizing from few samples remains a challenge.

Figure~\ref{fig:modalities} also displays the results of different modalities for each adverb-antonym pair. Different adverbs benefit from different modalities. For instance, `gently' and `firmly' are better distinguished with flow features, while recognition of `carefully' vs. `carelessly' benefits from the RGB modality. Overall the results of the different modalities are quite comparable, with RGB achieving 63.7, Flow 64.5 and the combination of both 63.9. The fusion of RGB and Flow not always being beneficial highlights that better fusion of modalities is needed for adverbs. This is a challenge since not only are different modalities useful for different adverbs, but the adverbs appearance is highly dependent on the action to which is applies and different actions are also better recognized with different modalities. 

\section{Adaptive Thresholding Hyperparameters}
\begin{figure}
    \centering
    \includegraphics[width=0.95\linewidth]{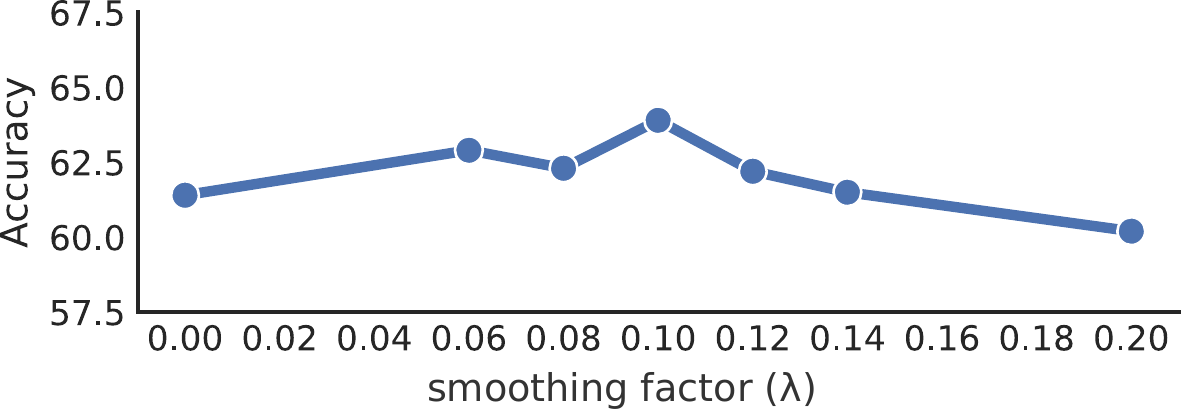}
    \vspace{-0.8em}
    \caption{Effect of smoothing factor $\lambda$.}
    \label{fig:smoothing}
    \vspace{-0.6em}
\end{figure}

\vspace{-0.3em}
\noindent\textbf{Effect of $\lambda$. }
In Fig.~\ref{fig:smoothing} we show the effect of smoothing factor $\lambda$ used in the adaptive thresholding (Eq.~\ref{eq:adaptive_threshold} in the main paper). The parameter determines how much the adverb-specific thresholds adapt and thus how much the model focuses on underrepresented adverbs. When $\lambda{=}0$ the original threshold $\tau$ is used for all adverbs. Fig.~\ref{fig:smoothing} shows that best results are obtained with $\lambda{=}0.1$, although any value $0.04\leq\lambda\leq0.14$ improves results.

\begin{figure}
    \centering
    \includegraphics[width=0.95\linewidth]{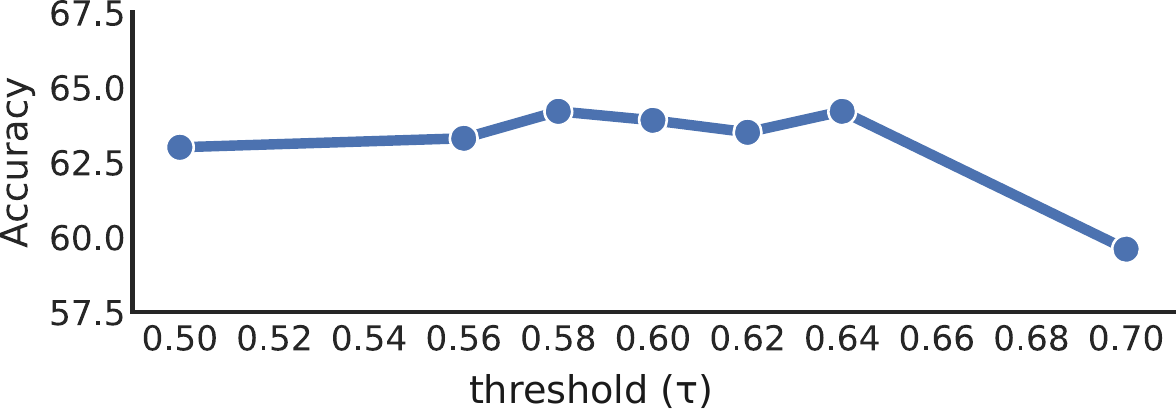}
    \vspace{-0.8em}
    \caption{Effect of base threshold $\tau$.}
    \label{fig:threshold}
    \vspace{-0.5em}
\end{figure}

\smallskip
\noindent\textbf{Effect of $\tau$}
The effect of base threshold for the pseudo-labeled adverbs used is shown in Fig.~\ref{fig:threshold}. The model is relatively insensitive to this parameter with any value $0.5\leq\tau\leq0.64$ being suitable.

\section{Long-Tail Results}

\begin{table*}[]
    \centering
    \resizebox{0.75\linewidth}{!}{\begin{tabular}{lcccccccccccccc}
    \toprule
         & && \multicolumn{3}{c}{Adverbs} && \multicolumn{3}{c}{Actions} & & \multicolumn{3}{c}{Pairs}\\
         \cmidrule{4-6} \cmidrule{8-10} \cmidrule{12-14}
         Method & All && Head & Mid & Tail & & Head & Mid & Tail & & Head & Mid & Tail\\
         \midrule
         Supervised only & 60.3 && 68.6	& 69.0	& 50.9	&& 62.3	& 56.5 &	57.3 && 78.8 & 56.5 & 57.3\\
         Pseudo-Label & 60.4 && 65.5 & 65.0 & 55.1 && 63.1 & 54.8 & 63.0	&& 78.2 & 54.8 & 62.3\\
         FixMatch & 61.2 && 65.8 & 63.4 & 57.4 && 60.5 & 63.9 & 55.0 && 79.2 & 63.9 & 55.0\\
         TCL & 58.3 && 64.9 & 74.6 & 55.3 && 66.5 & 56.5 & 58.7 && 77.8 & 56.5 & 58.7\\
         \midrule
         Ours & 63.9 && 68.9 & 69.1 & 58.2 && 65.3 & 60.2 & 59.7 && 84.2 &	60.3 & 59.7\\
         \midrule
    \end{tabular}}
    \caption{Results of the long-tail. We show adverb recognition results over the long-tails of adverbs, actions and their pairs. We split each into three categories of head, middle and tail. Our model successfully combats the long-tail of adverbs by increasing tail results over the supervised only baseline and maintaining a similar on the adverbs in the head and middle of the distribution.}
    \label{tab:long_tail}
\end{table*}

In Table~\ref{tab:long_tail} we investigate the ability of our method to recognize adverbs in the long tail distributions of adverb, actions and their compositions. For each we report adverb recognition results for the head, middle and tail of the distributions. For adverbs the head is defined as ${>}500$ instances and the tail is ${<}100$ instances. For actions the head is ${>}100$ instances and ${<}20$ instances. For the action-adverb pairs the head and tail are ${>}50$ and ${<}10$ instances respectively.

Table~\ref{tab:long_tail} shows the results on the 5\% split of VATEX Adverbs. We can see that our method improves results of the tail adverbs significantly over the supervised only baseline (50.9 to 58.2) while maintaining a similar performance for the head and middle of the adverb distribution. Other methods have a smaller improvement over the adverb tail while causing a decrease in performance at the head and middle of the distribution. For the action and pair distributions, our method increases results over the supervised only baseline for the head, middle and tail of the distributions. Other methods are better at certain parts of the distribution, for instance FixMatch obtains best performance at the middle of the action and pair distribution. However, their overall improvement is lower. From these results we can conclude that the success of our method is due to its improvement on the long-tail of adverbs.

\section{Baseline Implementation Details}
\vspace{-0.3em}
For all models we use the same backbone, video-text embedding functions ($f$ and $g$) and loss functions ($L_{act}$ and $L_{adv}$) as in our proposed method. The shared hyper-parameters are also common between our method and baselines. We outline the specific details for each below.

\noindent\textbf{Pseudo Label~\cite{lee2013pseudo}. } For this baseline we take the adverb in the closest embedded action-adverb text representation to be the hard pseudo-label, as defined in Eq.~\ref{eq:singlelabel}. All of these pseudo-labels are used in training, this approach does not use thresholding. The action-only labeled data and adverb pseudo-labeling is introduced at epoch 300, up until then the model is trained with the supervised data only. The loss functions for supervised and pseudo-labeled action-only data are given equal weighting. We experimented with different weightings and introducing the pseudo-labeling at different epochs as in~\cite{lee2013pseudo}, but empirically found these settings to perform best. This baseline uses both RGB and Optical flow modalities.

\noindent\textbf{FixMatch~\cite{sohn2020fixmatch}. } Fixmatch also uses hard pseudo-labels. An action-only video is first weakly augmented and the adverb in the closest embedded action-adverb text representation is taken as the pseudo-label (Eq.~\ref{eq:singlelabel}). A strongly augmented version of this video is then trained to predict this pseudo-label. We use the same augmentations as in the original paper~\cite{sohn2020fixmatch}. The weak augmentations are randomly flipping the video with 50\% probably and randomly translating the video by up to 12.5\% vertically and horizontally. The strong augmentations are those used in RandAugment~\cite{cubuk2020randaugment}: autocontrast, adjusting brightness, adjusting color balance, adjusting contrast, equalizing the video frame histogram, the identity function, posterizing, rotating, adjusting the sharpness, shearing along the x or y-axis, solarizing and translating the video along the x or y-axis. We randomly select two augmentations for each video in a batch, each with random magnitudes. Full details can be found here~\cite{sohn2020fixmatch}. The same augmentation with the same parameters are applied to all frames in a video. Fixmatch uses fixed thresholding where we use a threshold of $\tau=0.6$. This baseline uses the RGB modality.

\noindent\textbf{TCL~\cite{singh2021semi}. } TCL optimizes the consistency in predictions between a normal video and an augmented version played at twice the speed.  The agreement is maximized through an instance contrastive loss which encourages the two speeds of the video to have the same prediction for all adverbs. This is done with class logits in the original paper, since we use a video-text embedding space to learn adverbs we use the distance to each of the action-adverb compositions with the ground-truth action. There is also a group contrastive loss which optimizes agreement between the average predictions for groups of videos with the same pseudo-label. This grouping is done with the hard pseudo-label predicted from Eq.~\ref{eq:singlelabel}. As in the original paper we use a weighting of 9 for the instance contrastive loss and 1 for the group contrastive loss. This baseline uses the RGB modality.

\section{Dataset Licensing}
\vspace{-0.3em}
    Our adverb newly proposed adverb datasets are based on three existing video-text datasets: VATEX~\cite{wang2019vatex}, MSR-VTT~\cite{xu2016msr} and ActivityNet Captions~\cite{krishna2017dense}. All three of these datasets use videos from YouTube, as such all of the videos in these datasets and our adverb datasets use either the YouTube Standard License\footnote{\url{https://www.youtube.com/static?template=terms}} or the Creative Commons BY License\footnote{\url{https://creativecommons.org/licenses/by/3.0/us/}}.

\begin{figure}

    \centering
\includegraphics[width=\linewidth]{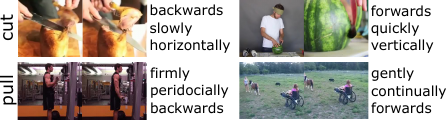}
    \vspace{-1em}
    \caption{Predictions of different adverbs for the same action.}
    \vspace{-1.2em}
    \label{fig:same_act}
\end{figure}

\section{Further Qualitative Results}
Fig.~\ref{fig:same_act} shows results of our model predicting different adverbs for an action. This demonstrates how adverbs could be applicable to anomaly detection in videos. For instance,the
figure shows ‘cut horizontally’, which we know is anomalous as ‘vertically’ is predicted for most ‘cut’ actions. 

\section{Action-Adverb Distribution}
\begin{figure}
    \centering
    \includegraphics[width=\textheight, angle=270]{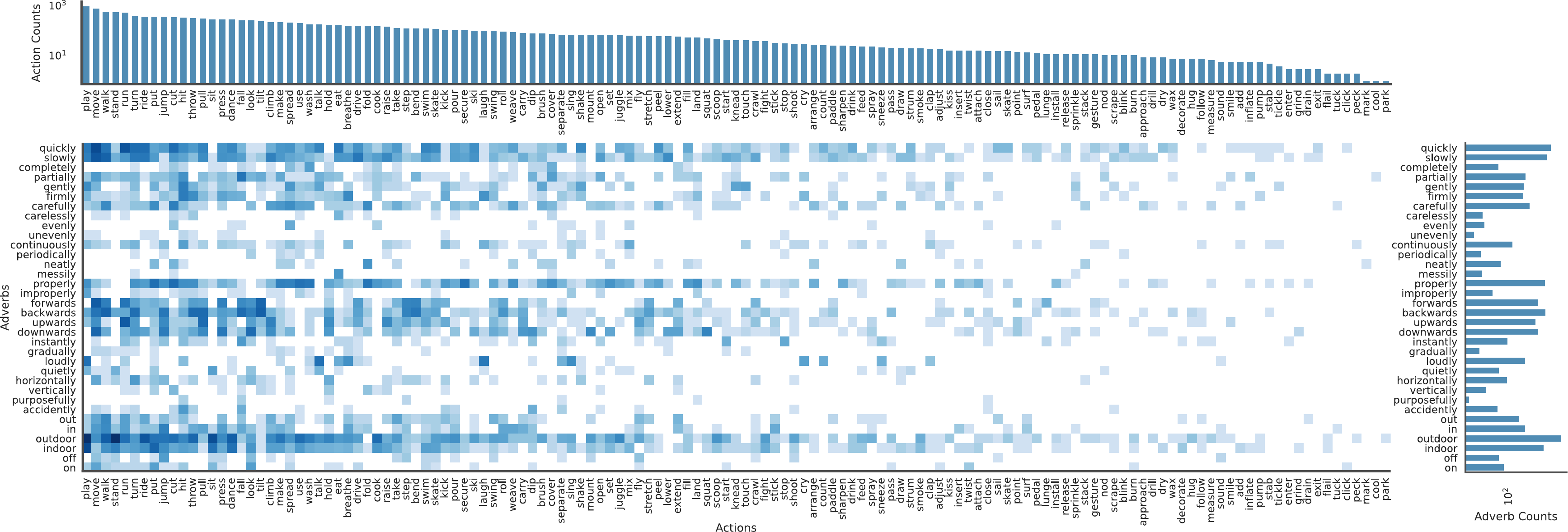}
    \caption{Distribution of action-adverb pairs in VATEX shown on a log-scale.}
    \vspace{-0.3em}
    \label{fig:vatex_dist_full}
\end{figure}

\vspace{-0.3em}
A full size version of the action-adverb distribution from the main paper is shown in Fig.~\ref{fig:vatex_dist_full}. Not only are the individual distributions of actions and adverbs long-tailed, but the action-adverb compositions are also heavily long-tailed.

\section{Supplementary Video}
\vspace{-0.3em}
We show video versions from Fig.~\ref{fig:qual_res} and Fig.~\ref{fig:bonus_task} from the main paper in the supplementary video.

\end{document}